\documentclass[letterpaper]{article} 
\usepackage{aaai24}  
\usepackage{times}  
\usepackage{helvet}  
\usepackage{courier}  
\usepackage[hyphens]{url}  
\usepackage{graphicx} 
\usepackage{natbib}  
\usepackage{caption} 
\usepackage{amsmath}
\usepackage{amssymb}
\usepackage{array}
\usepackage{lipsum}
\usepackage{adjustbox}
\usepackage{algorithm}
\usepackage{algorithmic}

\usepackage{newfloat}
\usepackage{listings}
\DeclareCaptionStyle{ruled}{labelfont=normalfont,labelsep=colon,strut=off} 
\floatstyle{ruled}
\newfloat{listing}{tb}{lst}{}
\floatname{listing}{Listing}
\title{Terrain Diffusion Network: Climatic-Aware Terrain Generation with Geological Sketch Guidance}
\author {
    Zexin Hu\textsuperscript{\rm 1},
    Kun Hu\textsuperscript{\rm 1},
    Clinton Mo\textsuperscript{\rm 1},
    Lei Pan\textsuperscript{\rm 2},
    Zhiyong Wang\textsuperscript{\rm 1}
}
\affiliations {
    \textsuperscript{\rm 1}The University of Sydney\\
    \textsuperscript{\rm 2}Civil Aviation Flight University of China\\
    zehu4485@sydney.edu.au, kun.hu@sydney.edu.au, clmo6615@uni.sydney.edu.au, leipan.cafuc@hotmail.com, zhiyong.wang@sydney.edu.au
}
\date{August 2023}
\begin{document}

\maketitle

\begin{abstract}
Sketch-based terrain generation seeks to create realistic landscapes for virtual environments in various applications such as computer games, animation and virtual reality. Recently, deep learning based terrain generation has emerged, notably the ones based on generative adversarial networks (GAN). However, these methods often struggle to fulfill the requirements of flexible user control and maintain generative diversity for realistic terrain. Therefore, we propose a novel diffusion-based method, namely terrain diffusion network (TDN), which actively incorporates user guidance for enhanced controllability, taking into account terrain features like rivers, ridges, basins, and peaks. Instead of adhering to a conventional monolithic denoising process, which often compromises the fidelity of terrain details or the alignment with user control, a multi-level denoising scheme is proposed to generate more realistic terrains by taking into account fine-grained details, particularly those related to climatic patterns influenced by erosion and tectonic activities. Specifically, three terrain synthesisers are designed for structural, intermediate, and fine-grained level denoising purposes, which allow each synthesiser concentrate on a distinct terrain aspect. Moreover, to maximise the efficiency of our TDN, we further introduce terrain and sketch latent spaces for the synthesizers with pre-trained terrain autoencoders. Comprehensive experiments on a new dataset constructed from NASA Topology Images clearly demonstrate the effectiveness of our proposed method, achieving the state-of-the-art performance. Our code and dataset will be publicly available.
\end{abstract}


\section{Introduction}
\par In the real world, terrains are subject to a variety of climatic conditions, such as temperature variations, erosion from water or wind, and the presence of vegetation. When employing an automated terrain generation method, it is essential to accurately depict such weather events and natural phenomena, while closely adhering to a user's structural guidance for controllability to avoid unintended outcomes. Yet, conventional controllable example-based and sketch-based methods \cite{talgorn2018real, guerin2017interactive} often compromise the generation of realistic terrains in terms of geologically precise fine-grained characteristics. 
For optimal fidelity, it requires more advanced controllable terrain synthesis techniques that are capable of producing diverse and intricate features with fine granularity.
\begin{figure}[h]
\centering
\includegraphics[width=0.39\paperwidth]{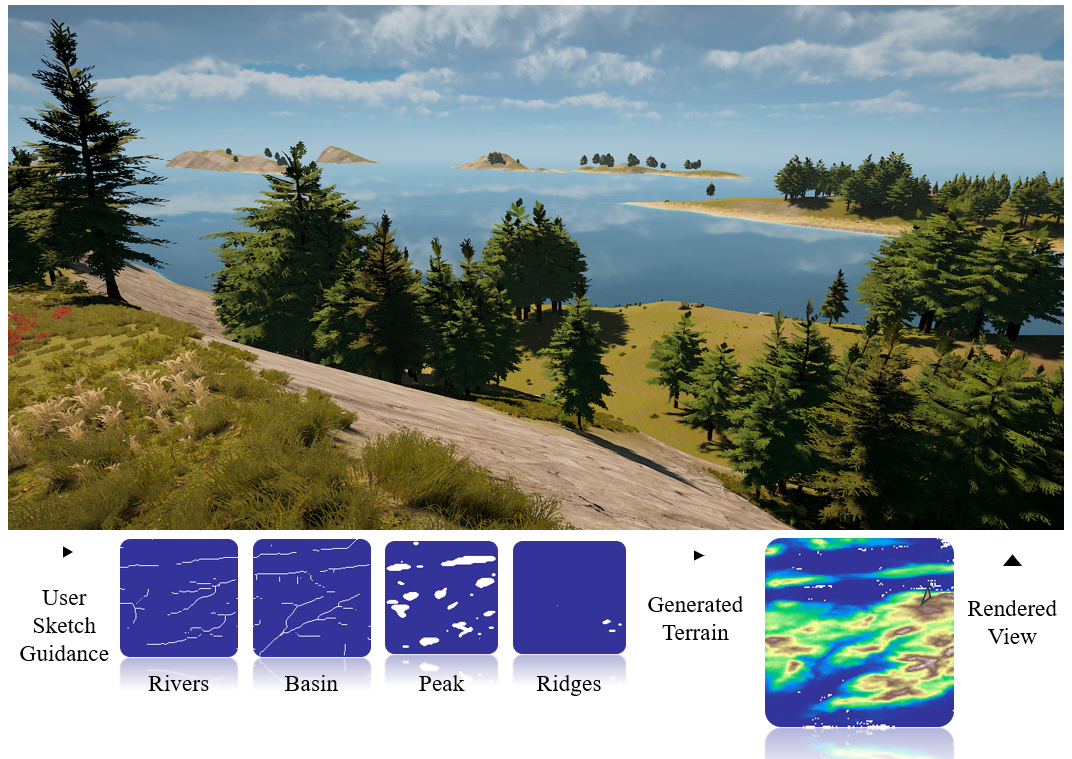}
\caption{Sample terrain generated by our Terrain Diffusion Network (TDN), which is controlled by user-provided sketch guidance with ridges, rivers, basins and peaks.}
\label{fig:teaser}
\end{figure}
\par Due to the great success of deep learning techniques for computer vision \cite{wu2016learning,jin2017towards,siarohin2018deformable,zheng2020p}, the potential of deep-learning methods has been explored for high fidelity terrain generation. The majority of these studies are based on generative adversarial networks (GAN) \cite{beckham2017step, spick2019realistic}. 
However, GANs still suffer the from trade off between fidelity and divergence to user-provided conditions.
Recently, diffusion methods \cite{dhariwal2021diffusion} have shown promising results in generating a wide variety of images based on user prompts. This suggests significant potential to overcome the challenges found in terrain generation: they foster diversity at a fine-grained level and offer the opportunity to enable the generation of climatic morphogenesis. 
Nonetheless, it comes with a significant challenge due to diffusion model's inherent stochastic nature against the user's capability to guide the terrain generation process. 
To the best of our knowledge, there is no existing study with the controllable diffusion architecture for fine-grained and diversified terrain generation. 

\par Therefore, in this paper, we propose a novel deep learning method based on the diffusion approach, namely terrain diffusion network (TDN). It employs a generation process by denoising Gaussian noise-perturbed terrain latent representations to produce realistic and visually appealing terrains that closely align with the user provided sketch guidance. 
TDN allows users to explicitly define the structural-level terrain characteristics through their input sketches for rivers, ridges, basins, and peaks, as shown in Fig. \ref{fig:teaser}. To adequately consider the user guidance for terrain generation, a multi-level denoising scheme with a terrain sketch encoder, that aims to structural consistency, is devised to replace the monolithic denoising in the conventional diffusion process. It employs a coarse-to-fine strategy, utilising multiple terrain synthesizers with direct user guidance to articulate structural, intermediate, and fine-grained geological patterns. This facilitates the transformation of low-resolution inputs into high-resolution terrain outputs. Specifically, different denoisor weights are learnt for distinct synthesis stages, and they can focus on different levels of terrain patterns. In detail, the structural synthesiser focuses more on the overall terrain components, such as rivers, ridges, basins, and peaks, whilst the fine-grained synthesiser provides further terrain details such as climatic patterns including geomorphological erosion and tectonic events. 
To facilitate the efficiency of our TDN, we further introduce terrain and sketch latent spaces with lower dimensions by utilizing pre-trained autoencoders. 

In summary, the key contributions of this study are:
\begin{itemize}
    \item A novel deep learning based approach that takes multiple geometric factors into account for controllable terrain generation. 

    \item A multi-level denoising synthesizer to formulate both structural and fine-grained terrain patterns aiming for producing realistic and climatic-aware terrain patterns. 

    \item A new dataset is constructed from NASA Topology Images \cite{Allen_2005} to evaluate the effectiveness of the proposed TDN. Comprehensive experiments demonstrate that the proposed TDN is able to achieve high-quality realistic terrain synthesis with flexible user controls.
\end{itemize}

\section{Related Work}

\subsection{Procedural Terrain Generation}
\par Procedural terrain generation was first introduced by \citet{mandelbrot1982fractal}. Generally, procedural methods involved manipulating fractal noise by using a predefined set of rules, algorithms, or functions or input parameters to mimic visually faithful terrain features. Over time the methods became increasingly computationally efficient due to various research efforts that had been made. 
While procedural models were usually computationally efficient, they were limited due to the lack of the capability to involve user control. This issue stemmed from the stochastic nature of the approach. To address this limitation, constrained fractals had been introduced, which combined deterministic features such as user-prescribed terrain projections \cite{belhadj2005modeling} or deterministic splines \cite{derzapf2011river} with stochastic fractals. More recently, \citet{10.1145/2461912.2461996,gaillard2019dendry} both proposed pipelines that allow users to have more flexibility for controllable generation by making minor adjustments to input parameters. However, while providing user control, terrains generated with procedural methods tend to appear pristine and lack signs of erosion and weathering, making them less realistic than real ones shaped by climatic morphogenesis.

\subsection{Simulation-based Terrain Generation}

\par Simulation-based methods addressed some of the trade-off challenges experienced by procedural methods. They typically emulated geomorphological processes, such as erosion and tectonics, to generate realistic terrain features. 
Simulating erosion with approximately tuned parameters helped create more realistic terrain height maps \cite{cordonnier2016large}.
\citet{cordonnier2017sculpting} simulated tectonic features to generate ranges, valleys and other large-scale terrain features. \citet{krivstof2009hydraulic} simulated hydraulic events to generate terrains. Recent research took advantage of the advancements in computational capacity to enhance user interactions with terrain generation \cite{benes2001layered,mei2007fast,vanek2011large}.For instance, in \citet{cordonnier2017authoring} a simulation system was proposed to simulate geomorphological events at an unprecedentedly high speed. 
Nonetheless, simulation-based methods still suffered from the limited flexibility of user control and expensive computational costs for more extensive scenarios. 
\begin{figure*}[h]
\centering
\includegraphics[width=0.8\paperwidth]{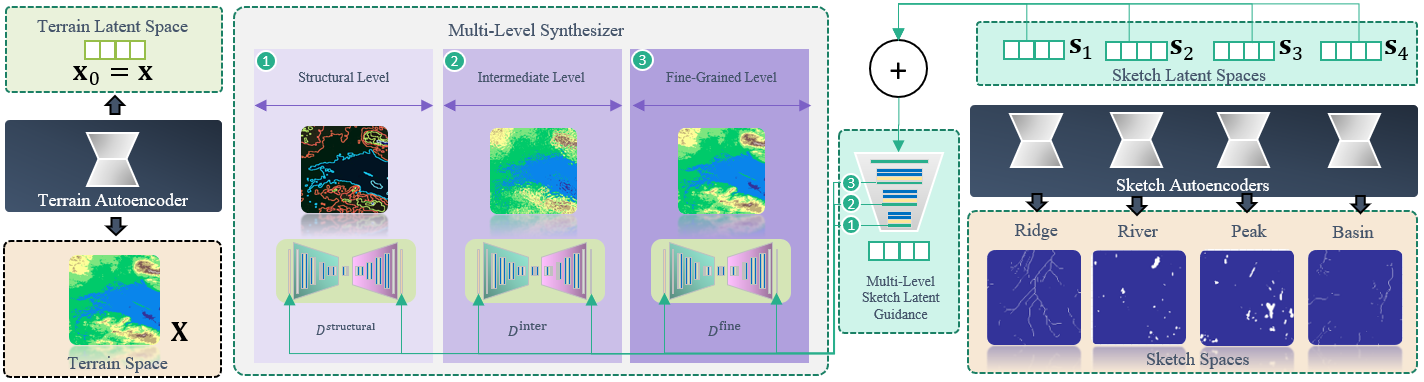}
\caption{Illustration of our proposed Terrain Diffusion Network (TDN).
}
\label{fig:figure2}
\end{figure*}

\subsection{Example-based Terrain Generation}

Procedural and simulation-based methods often required meticulous parameter tuning to generate desired terrains and lacked intuitive ways to involve user control. In contrast,
example-based methods used more intuitive user guidance such as sketches and images for terrain generation \cite{spick2019realistic, li2006example}. 
\citet{zhou2007terrain} proposed a more direct user control scheme that blends patches from real terrains containing height fields with user-defined sketches. 
\citet{vst2008interactive} incorporated small-scale simulation-based models with an interactive editing scheme. \citet{rusnell2009feature} utilized surface deformation to match user constraints. \citet{tasse2020first,scott2021example} generated new terrains based on existing samples. 
While example-based methods generally provided a high level of control, they often involve a trade-off between realism/geological correctness and  controllability \cite{hnaidi2010feature,gain2009terrain,vanek2011large}. In other words, the increased control can typically results in relatively lower-quality terrains. 

\subsection{Deep-Learning based Terrain Generation}
\par The deep learning approach had gained impressive performance and increasing popularity for the fields in computer vision and graphics. Specifically, deep generation methods such as Generative Adversarial Networks (GAN) \cite{radford2015unsupervised} and conditional GANs (cGAN) \cite{mirza2014conditional}, which accept multimodal guidance, such as images and texts, to be conditioned on the generation process, have been applied to sketch-based terrain generation, achieving state-of-the-art results \cite{guerin2017interactive}. However, due to the inherently stochastic nature of GANs, the generative process still lacks user control for fine-grained details without using multiple datasets~\cite{salimans2016improved}. In recent years, a new technique for a generation - the diffusion approach \cite{song2020denoising} has emerged and achieved superior performance compared to GANs \cite{dhariwal2021diffusion, rombach2022high, zhang2023adding, meng2021sdedit} in an extensive range of fields. The existing GAN-based terrain generation systems fall short in their ability to incorporate climate specific factors solely using real-world terrain data. To produce realistic terrain, it needs the training of several distinct GAN models, each one focused on a different dataset specifically targeted at distinct climatic events. 

\par 
Recently, a significant progress in the field of image-to-image generations has been achieved by leveraging diffusion models\cite{li2023gligen, huang2023composer, wang2023diffsketching,Mei_Patel_2023}. These methods deliver an unparalleled level of fidelity and user control\cite{zhang2023adding,Zhang_Zhao_Yu_Tian_2023}.
Although diffusion models demonstrated unprecedented diversity and fidelity in its generation, the inherent stochastic nature of diffusion process imposes significant challenges for implementing effective user control. On the contrary, our TDN leverages a terrain specific diffusion approach integrating multi-level user guidance to generate climate-aware and plausible terrain that align with user sketches. Our approach empowers users with granular control while eliminating the necessity for training on multiple specialized datasets.

\section{Methodology}

As shown in Fig. \ref{fig:figure2}, our proposed Terrain Diffusion Network (TDN) is composed of two primary components: terrain and sketch autoencoders and multi-level terrain synthesisers. The autoencoders with U-Net \cite{ronneberger2015u} like structures streamlines a terrain map by converting it into a lower-dimensional latent representation for reducing computational demands during the diffusion process. The diffusion process leverages multi-level synthesisers to create a latent terrain representation, with diversity and flexibility under the guidance of user-provided sketches.

\subsection{Controllable Terrain Synthesis}

\par Our method uses a set of user's sketches as the guidance, including rivers, ridges, basins, and peaks to generate a realistic terrain map \(\mathbf{X}\in \mathbb{R}^{C\times W\times H}\). We denote the sketch guidance as a set \(\mathbf{\mathbb{S}}=\{\mathbf{S}_1,...,\mathbf{S}_n,...,\mathbf{S}_N\}\), with each user sketch \(\mathbf{S}_{n} \in \mathbb{R}^{C\times W \times H}\). These sketches and the terrain map can be viewed as one channel (\(C=1\)) images. Fig. \ref{fig:teaser} depicts an example of paired \(\mathbf{X}\) and \(\mathbf{\mathbb{S}}\), where $N=4$ in our study covers four major control factors in terrain synthesis. 

\subsection{Terrain and Sketch Autoencoder}

\par A terrain autoencoder is introduced to compress the terrain data into a latent space.
This enables the projection of terrain data into a more manageable, lower-dimensional space,
thereby reducing the computational demands of the diffusion process \cite{rombach2022high}. The autoencoder is based on a U-Net architecture with an encoder \(Z\) and a decoder \(Z'\). In the encoding stage, the target terrain is funneled through the down-sampling encoder components to generate the latent representation. Subsequently, in the decoding stage, the terrain is reconstructed via the up-sampling decoder components.
The overall process can be formulated as \(\mathbf{X} \approx Z^{'}(\mathbf{x} = Z(\mathbf{X}))\), where \(\mathbf{x}\in\mathbb{R}^{d}\) is the latent representation of \(\mathbf{X}\). A perceptual loss function is adopted to train the autoencoder \cite{hinton2006reducing}, which encourages the autoencoder to preserve high-level features integral to the terrain perception. The computation of perceptual loss involves passing both the ground truth terrain and the reconstructed terrain through a pre-trained loss network. Mathematically, we have \cite{pan2016perceptual}:
\begin{equation}
\mathcal{L}(\mathbf{X},\mathbf{x}_{rec}) = \frac{1}{\mathbf{C}_{j}\mathbf{H}_{j}\mathbf{W}_{j}}\left \| \phi _j(\hat{\mathbf{X}}) - \phi_{j}(\mathbf{X})) \right \|.
\label{eq:1}
\end{equation}

\par 
Similarly, we adopt autoencoders with similar structures to encode user-provided sketch guidance. Encoder networks \(Z_n\), \(n=1,...,N\) are introduced for rivers, ridges, basins, and peaks, respectively, to compress \(\mathbf{S}_n\) into \(\mathbf{s}_n\) the \(n\)-th sketch latent space. 
An additional transformer is employed to integrate the information from these encoded latent data and create a comprehensive yet condensed overview of the input sketch guidance \(\mathbb{S}\), as illustrated in Fig. \ref{fig:teaser}. 
Subsequently, this aggregated sketch representation is processed through a convolutional layer and a sketch guidance vector can be obtained as \(\mathbf{s}\in\mathbb{R}^d\). 
Note that \(\mathbf{s}\) aligns in dimensionality with the terrain latent representation \(\mathbf{x}\), ensuring a coherent and efficient guided diffusion process.

\subsection{Terrain Diffusion}
\par The diffusion model conducts a procedure that involves adding noise to a sample in the latent terrain space and then using a deep neural network to reverse the noise-perturbed sample back to its original latent representation. 
Specifically, the diffusion process introduces a Gaussian noise to the latent representation iteratively. At the $t$-th step, the relationship between \(\mathbf{x}_{t}\) and \(\mathbf{x}_{t-1}\) can be formulated as: 
\begin{equation}
q(\mathbf{x}_{t}|\mathbf{x}_{t-1}) = \mathcal{N}(\mathbf{x}_{t};\mathbf{\mu}_{t} = \sqrt{1-\beta_{t}}\mathbf{x}_{t-1},\mathbf{\Sigma}_{t} = \beta_{t}\mathbf{I})),
\label{eq:2}
\end{equation}
where \(q(\mathbf{x}_{t}|\mathbf{x}_{t-1})\) is a conditional probability for a forward diffusion process, which follows a Gaussian distribution with mean \(\mathbf{\mu}_{t}\) and variance \(\mathbf{\Sigma}_{t}\). 
In practice, 
starting from the original terrain data \(\mathbf{x}_{0}=\mathbf{x}\), the forward diffusion process is traceable as follows
\cite{ho2020denoising}:
\begin{equation}
    q(\mathbf{x}_{1:T}|\mathbf{x}_0) = \prod_{t=1}^{T} q(\mathbf{x}_{t}|\mathbf{x}_{t-1}).
\label{eq:3}
\end{equation}
By reparametrising Eq. (\ref{eq:2}) with \(\epsilon_{0},...,\epsilon_{t-2},\epsilon_{t-1}\sim \mathcal{N}(0,I)\), and \(\alpha_{t} = 1 - \beta_{t}, \bar{\alpha}_{t} = \prod_{s=0}^{t} \alpha_{s}\), we have:
\begin{equation}
    \mathbf{x}_{t} \sim q(\mathbf{x}_{t}|\mathbf{x}_{0}) = \mathcal{N}(\mathbf{x}_{t};\sqrt{\bar{\mathbf{\alpha}_{t}}}\mathbf{x}_{0}, (1-\bar{\alpha_{t}})\mathbf{I}).
\label{eq:4}
\end{equation}

Now the reverse process can be conducted based on a distribution, of which the probability is defined as \(p(\mathbf{x}_{t-1}|\mathbf{x}_{t})\). 
It formulates a terrain latent representation in the previous sampling step $t-1$ by providing its current state at $t$. Specifically, this probability and thus the corresponding reverse process are estimated by a neural network $ D_{\theta}(\mathbf{x}_t|\theta)$, where $\theta$ contains learnable weights. Mathematically, we have: 
\begin{equation}
    p_\theta(\mathbf{x}_{t-1}|\mathbf{x}_{t}) = \mathcal{N}(\mathbf{x}_{t-1};\mu_{\theta}(\mathbf{x}_{t},\mathbf{t}),\Sigma_{\theta}(\mathbf{x}_{t},\mathbf{t})),
\label{eq:5}
\end{equation}
where $p_\theta(\mathbf{x}_{t-1}|\mathbf{x}_{t})$ is an estimation of $p(\mathbf{x}_{t-1}|\mathbf{x}_{t})$. Note \(D_{\theta}(\mathbf{x}_t|\theta)\) is also known as a denoiser or synthesiser in diffusion. 
The trajectory of the reverse process would be: 
\begin{equation}
    p_{\theta}(\mathbf{x}_{0:T}) = p_{\theta}(\mathbf{x}_{T})\prod_{t=1}^{T} p_{\theta}(\mathbf{x}_{t-1}|\mathbf{x}_{t}).
\label{eq:6}
\end{equation}

According to \citet{luo2022understanding},  \(\theta\) can be optimized with a loss:
\begin{equation}
    L_{t} = \mathbb{E}_{\mathbf{x}_{0},\mathbf{t},\mathbf{\epsilon}}[||\mathbf{\epsilon}-\mathbf{\epsilon}_{\theta}(\sqrt{\bar{\mathbf{\alpha}_{t}}}\mathbf{x}_{0}+\sqrt{1-\bar{\mathbf{\alpha}}_{t}}\mathbf{\epsilon},\mathbf{t})||^{2}],
\label{eq:7}
\end{equation}
where \(x_{0}\) is the noise free terrain latent representation, \(\epsilon\) is the noise sampled from \(\mathcal{N}(0,I)\) and the noise scheduler \(\bar{\alpha}_{t} = \prod_{s=0}^{t} \alpha_{s}\). Our primary goal is to utilize our neural network \(D_{\theta}\) to estimate the noise $\epsilon$ as $\epsilon_\theta$.

\subsection{Multi-Level Sketch Guidance Integration}
\par The terrain generation is challenging to control via an unconditional diffusion process. Thus, we propose a diffusion process uses sketch guidance for enhanced controllability in terrain synthesis. Specifically, by giving the sketch prompt \(\mathbb{S}\) and its latent representation \(\mathbf{s}\) from users, a conditioned diffusion process can be formulated as \cite{song2020score}:  
\begin{equation}
    p_{\theta}(\mathbf{x}_{0:T}|\mathbf{s}) = p_{\theta}(\mathbf{x}_{T})\prod_{t=1}^{T} p_{\theta}(\mathbf{x}_{t-1}|\mathbf{x}_{t},\mathbf{s}).
\end{equation}

It conditions on Eq. (\ref{eq:6}) with added user guidance \(\mathbf{s}\).
By formulating Eq. (\ref{eq:7}) with the condition of user sketch guidance \(\mathbf{s}\), the conditioned synthesiser can be optimized regarding its weights \(\theta\) by solving the following optimization problem: 
\begin{equation}
    L_{t} = \mathbb{E}_{\mathbf{x}_{0}|s,\mathbf{t},\mathbf{\epsilon}}[||\mathbf{\epsilon}-\mathbf{\epsilon}_{\theta}(\sqrt{\bar{\mathbf{\alpha}_{t}}}\mathbf{x}_{0}|s+\sqrt{1-\bar{\mathbf{\alpha}_{t}}}\mathbf{\epsilon},\mathbf{t})||^{2}].
\label{eq:9}
\end{equation}

\par In practice, how to provide appropriate guidance remains as an open question. While the conventional classifier-free \cite{ho2022classifier} or the cross-attention mechanisms used in methods such as stable diffusion \cite{rombach2022high} may work in some contexts, they are not the optimal solution for terrain generation as they were primarily designed for dimensionality matching (e.g., caused by cross-modal discrepancy) between their guidance and target data. Unlike these approaches, our terrain sketch encoder aims to preserve the structural consistency between the sketch guidance and the terrain latent representations. We introduce a mechanism to integrate sketch guidance directly into the denoising process, where the guidance $\mathbf{s}$ shares identical dimension $d$ with the noise perturbed terrain representation $\mathbf{x}_t$. 
Specifically, \(\mathbf{s}\) is concatenated with \(\mathbf{x}_{t}\) natively through a convolution layer. It is worth noting \(\mathbf{s}\) is constant throughout all diffusion steps \(t\). To this end, the network is able to estimate \(\hat{\mathbf{x}}_{0}|\mathbf{s}\) in an iterative manner, which is the generated output. We also enhance the control by employing skip connection, allowing the generated terrain be optimised towards user guidance in the cases where diffusion process introduce too much diversity. To summarize, this intrinsic integration enables the diffusion process to retain structural correspondence in line with the user provided sketches.

\subsection{Multi-Level Synthesisers}
\par Existing methods primarily conduct the diffusion process iteratively by applying a single denoiser neural network. The weights of the denoiser are fixed and incorporate the guidance similarly for all reconstruction steps. Thus, the user's sketch guidance is applied with the same weight irrespective of the current noise level at a particular step \(t\), forcing the diffusion process to balance between adhering to user's sketch and generating properable terrain
By dealing with both structural level terrain patterns and fine-grained level terrain climatic characteristics, a single synthesiser with the conventional approach would become a generalist that lacks specialization. This falls short of precisely generating fine-grained details or obtaining consistency in adhering to the user's structural sketch guidance. 
Therefore, in our approach, we propose a diffusion process within multiple latent spaces, where the diffusion process is guided by multiple synthesisers of stratified geological levels: structural-level, intermediate-level, and fine-grained level. 
Specifically, the fine-grained synthesiser plays a role in preserving and reconstructing climatic details, whilst at the structural-level synthesiser generates terrain patterns that closely match the user's input sketch guidance.

\par Unlike e-Diff \cite{balaji2022ediffi}, our multi-synthesiser employs different latent spaces to focus on different granular levels of the terrain. \citet{leeb2022exploring} showed that varying latent space dimensions affects different aspects of terrain generation. We crafted our synthesizer with different latent spaces, each tailored to specific terrain generation levels. User sketches are encoded to the terrain's latent dimension for direct diffusion integration.

\par The structural-level synthesiser controls the generation of lowest resolution latent representation. At this stage, the primary structural information is expected to be reconstructed. Specifically, based on Eq. (\ref{eq:9}), we optimize \(\theta_{structural}\) with: 
\begin{equation}
    L^\text{structural}_{t} = \mathbb{E}_{\mathbf{x}_{0}|s,\mathbf{t},\mathbf{\epsilon}}[||\mathbf{\epsilon}-\mathbf{\epsilon}_{\theta_{s}}(\sqrt{\bar{\mathbf{\alpha}}_{t}}\mathbf{x}_{0}|s+\sqrt{1-\bar{\mathbf{\alpha}}_{t}}\mathbf{\epsilon},\mathbf{t})||^{2}].
\end{equation}

Likewise, we further introduce an intermediate synthesiser \(D_{inter}^{\text{inter}}\allowbreak (\mathbf{x}_t|\theta_{inter})\). It controls the generation of the coarse level details and plays a balancing role between reconstructing both structural and fine-grained level perturbed patterns, where $\theta_{inter}$ is optimized by:
\begin{equation}
    L^\text{inter}_{t} = \mathbb{E}_{\mathbf{x}_{0}|s,\mathbf{t},\mathbf{\epsilon}}[||\mathbf{\epsilon}-\mathbf{\epsilon}_{\theta_{i}}(\sqrt{\bar{\mathbf{\alpha}}_{t}}\mathbf{x}_{0}|s+\sqrt{1-\bar{\mathbf{\alpha}}_{t}}\mathbf{\epsilon},\mathbf{t})||^{2}].
\end{equation}
Lastly, the fine-grained synthesiser \(D_{fine-grained}(\mathbf{x}_t|\theta_{f})\) controls the generation of the fine grained details by focusing on denoising terrain details such as climatic factors to synthesis results that closely match with realistic terrains. 
It is optimized with:
\begin{equation}
    L^\text{fine-grained}_{t} = \mathbb{E}_{\mathbf{x}_{0}|s,\mathbf{t},\mathbf{\epsilon}}[||\mathbf{\epsilon}-\mathbf{\epsilon}_{\theta_{f}}(\sqrt{\bar{\mathbf{\alpha}}_{t}}\mathbf{x}_{0}|s+\sqrt{1-\bar{\mathbf{\alpha}}_{t}}\mathbf{\epsilon},\mathbf{t})||^{2}].
\end{equation}
Despite similar structures, the three synthesizers are trained separately. Their varied weights allow for tailored guidance at different stages of the diffusion process.

\section{Experiments \& Discussions}

\subsection{Dataset}
\par The dataset used in this study is collected from NASA and the key features of terrains are extracted using Pysheds \cite{bartos_2020} packages. The images have a 1:6400 meter scale, with each extracted elevation map being \(144\times 144\). The sketches extracted from an image contain basins, peaks, rivers, and ridges. Sketches are produced using Pysheds' Digital Elevation Map (DEM) conditioning techniques, such as pit filling and flow direction determination. We obtained 10,446 samples for training and for 2,611 testing.
\subsection{Implementation Details}
\par \par Our model generates a latent representation of a terrain that is initially perturbed with noise through a specialized noise scheduler. To ensure a more seamless and gradual process of noise addition, we incorporate a cosine scheduler originally proposed by \citet{nichol2021improved}.
The synthesisers in our TDN are designed with U-Net like architectures, each of which comprises three downsampling blocks, three upsampling blocks, and a middle block with the self-attention mechanism employed for embedding the diffusion step \(t\) that has 8 heads. With a total of approximately 1.216 billion parameters, the model's training is conducted with a learning rate of 1.0e-05 and a batch size of 6. 
During the inference stage, TDN takes a set of user sketches as the input and iteratively generates a noise-free terrain latent representation. In total, 36 steps are taken to derive the final latent representation. 
An Nvidia RTX 3090 GPU is adopted for experiments. 

\begin{figure*}[h]
\centering
\includegraphics[width=\textwidth]{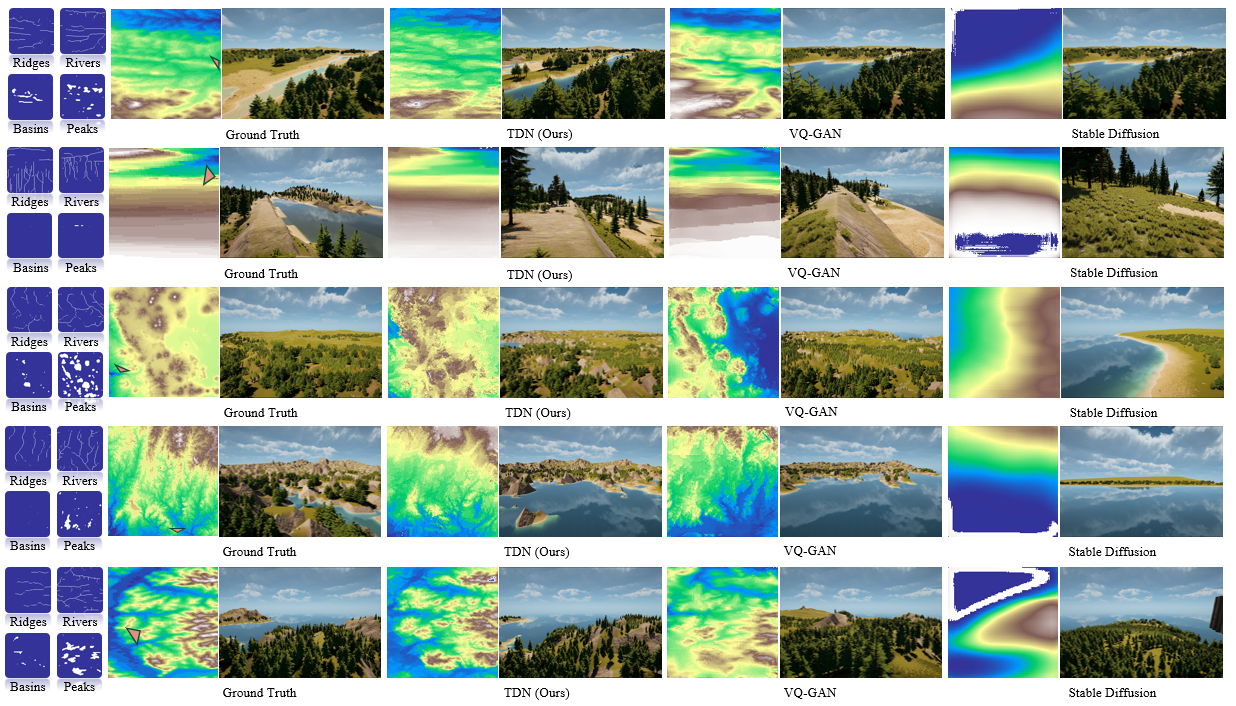}
\caption{Comparisons between various terrain generation methods with user input sketches, height map and rendered view. }
\label{fig:figure3}
\end{figure*}

\subsection{Overall Performance}
\par To demonstrate the effectiveness of our proposed TDN for terrain generation, we compare it with two state-of-the-art approaches: 1) GAN-based methods: GAN \cite{guerin2017interactive} and VQGAN \cite{esser2021taming}, and 2) diffusion-based methods: Stable Diffusion \cite{rombach2022high} and ControlNet \cite{zhang2023adding}, Gligen \cite{li2023gligen}. We utilize the metrics: Frechet Inception Distance (FID) \cite{chong2020effectively} and Mean Squared Error (MSE) to measure the performance of these methods by assessing the similarity between the generated terrains and their corresponding ground truth.

\begin{table}[h!]
\begin{center}
\begin{adjustbox}{width=0.47\textwidth}
\begin{tabular}{|p{5cm}|p{1.5cm}<{\centering}|p{1.5cm}<{\centering}|}
 \hline
Methods & FID $\downarrow$ & MSE $\downarrow$ \\
 \hline\hline
 GAN (\citeyear{guerin2017interactive})  & 4.65986 & 0.03910\\
 VQGAN (\citeyear{esser2021taming}) &   6.51174  & 0.05481\\ \hline
 Stable Diffusion (\citeyear{rombach2022high}) & 8.23260 & 0.08540\\
 ControlNet (\citeyear{zhang2023adding}) & 7.89227 & 0.08293\\
 GliGen (\citeyear{li2023gligen}) & 7.15267 & 0.07106\\
 \hline\hline
 TDN (Ours) &  \textbf{0.44023} & \textbf{0.00590}\\
 \hline
\end{tabular}
\end{adjustbox}
\caption{Comparisons between terrain generation methods. }
\label{table:1}
\end{center}
\end{table}

\par Our proposed TDN exhibits superior performance and surpasses all existing techniques in quantitative evaluations, as shown in Table 1. The FID scores suggest that TDN is able to generate terrain with high fidelity. 
Note that general diffusion methods including Stable Diffusion, ControlNet and Gligen are inferior to GAN-based methods, especially regarding the structural patterns as indicated by the FID metric. This is primarily due to diffusion model without specific terrain domain design struggles to extract appropriate features. Stable Diffusion is not able to accurately condition on multiple sketches, and causing the subsequent poor performance of ControlNet and Gligen as they relied on the pre-trained stable diffusion model.
In contrast, our TDN successfully addresses this issue with terrain specific guidance incorporated in our diffusion process, which demonstrates the mechanism's effectiveness for terrain specific tasks.
Our TDN has an edge, compared to GAN-based methods, as GAN is unable to provide adequate adherence of user sketech data whilst maintaining a high perceptual quality. Although VQGAN has gained superior performance for extensive scenarios, vanilla GAN outperforms VQGAN for terrain synthesis.
GAN outperformed VQ-GAN due to different mechanisms used for integrating the user sketches: GAN adopts a straightforward way to involve guidance via convolution, whilst VQGAN utilizes cross-attentions for this purpose. It indicates the uniqueness of the terrain generation, which requires distinct mechanisms to ensure the synthesis quality.
\par Fig. \ref{fig:figure3} visualises generated terrains, comparing our TDN method with GAN, Stable Diffusion (SD), and the ground truth. GAN and SD were selected as they are top performers of their respective approaches. Our TDN exhibits the ability to preserve a compelling degree of terrain's natural fidelity while adhering to user input sketch. This ability allows the generation of a more climate-aware and plausible terrain compare to GAN-based methods. The visual results align with our quantitative assessments, the distribution of the terrain generated by our model exhibits a closer representation of the actual terrain data, as compared to those produced by GAN and Stable Diffusion. Specifically, the Stable Diffusion model is unable to generate a closely matched terrain with the conventional cross-attention mechanism to integrate guidance, whilst our multiple synthesisers are capable of individually learning the denoising process at each step and significantly improves the control and performance that is specific to terrain generation. 

\subsection{Ablation Study}
\par Ablation studies are conducted to further evaluate the effectiveness of individual mechanisms in TDN for terrain generation. The quantitative and qualitative results are shown in Table 2 and Fig. 4, respectively. 

    \subsubsection{TDN w/o Multi-Level Synthesizers} This setting follows a conventional diffusion design with a single denoiser. FID and MSE drop 13.18\% and 16.55\% compared with TDN's multi-level synthesizer scheme, respectively. TDN's multi-level scheme introduces a degree of flexibility in the weighting of guidance, offering enhanced control. This dynamic guidance enables TDN to better adapt to various user demands during generation. 
    
    \subsubsection{TDN w/o Intermediate-Level Synthesizers} 
\begin{figure}[h]
\centering
\includegraphics[width=0.5\textwidth]{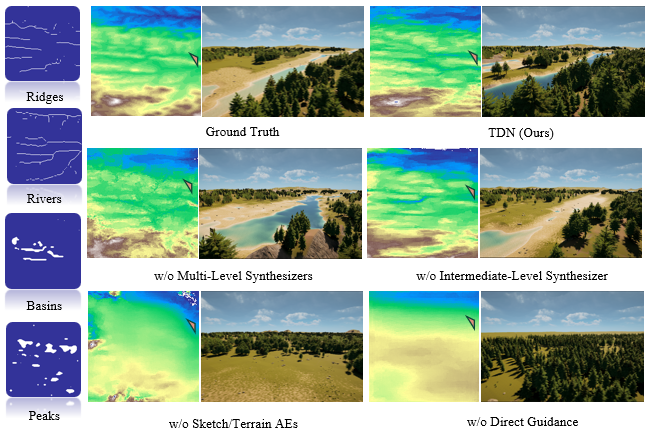}
\caption{Qualitative study on various mechanisms.}
\label{fig:figure4}
\end{figure}

This setting removes the intermediate-level synthesizer of TDN, and keeps the structural and fine-grained levels as a two-level denoising scheme. This setting significantly compromises the generation, compared with its three-level counterpart (TDN). It indicates the necessity to connect and transit from the structural stage to the fine-grained stage. The setting without the intermediate-level synthesizer suffers on the structual level generation as well as fine-grained generation. In Fig. \ref{fig:figure4}, it loses some of its fine-grained details, and clearly deviates from the ground truth.
    Intriguingly, our experiments show that a model with a single synthesizer can sometimes surpass the performance of  utilizing two synthesizers. This suggests that an efficient performance isn't merely a function of the number of synthesizers, but also lies in the nuances of how the individual synthesizer is employed. Such findings open avenues for further exploration into the specific roles and optimal usage of synthesizers in the terrain generation process.

    \subsubsection{TDN w/o Sketch Autoencoders (AEs)} This setting removes the sketch autoencoders, and the sketches share the autoeconder with  terrains. The sketch autoencoders facilitate the learning of suitable weights for various sketch inputs. Fig. \ref{fig:figure4} illustrates the effectiveness of the sketch autoencoders in our terrain generation process, which 
    adheres more closely to the sketch.

    \subsubsection{TDN w/o Direct Sketch Guidance}

    This setting follows a general practice that provides the integration of the guidance with the terrain via a cross-attention mechanism. It can be observed that our direct integration strategy is more effective for the consistency between the synthesized terrain and the input sketches, taking advantage of their directly matched latent patterns. Stable Diffusion and ControlNet have same issues as the control provided through cross-attention had an adverse effect for terrain generation.

\begin{table}[h!]
\begin{center}
\begin{adjustbox}{width=0.47\textwidth}
\begin{tabular}{ |p{5cm}|p{1.5cm}<{\centering}|p{1.5cm}<{\centering}|  }
 \hline
Methods & FID $\downarrow$ & MSE $\downarrow$ \\
 \hline\hline
 TDN (Ours) &  \textbf{0.44023} &  \textbf{0.00590}\\
 \hline\hline
 w/o Multi-Level Synthesisers   & 0.50707  &  0.00707\\
 w/o Intermediate-Level Synthesizer &   1.88794  &  0.01769\\
 w/o Terrain \& Sketch AEs & 1.94008 & 0.03118\\
 w/o Direct Sketch Guidance & 1.89567 &  0.01950\\
 \hline
\end{tabular}
\end{adjustbox}
\caption{Ablation study on the proposed TDN.}
\label{table:2}
\end{center}
\end{table}

\subsection{Out-of-Domain Generation}

\par
\begin{figure}[h]
\centering
\includegraphics[width=0.5\textwidth]{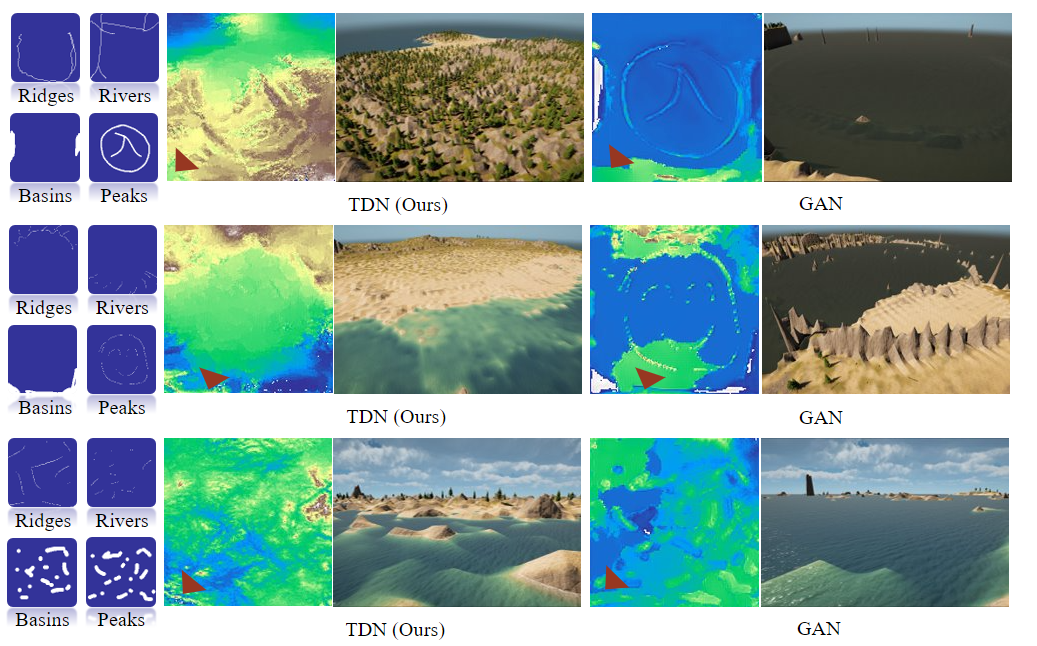}
\caption{Out-of-domain terrain generation.}
\label{fig:figure5}
\end{figure}

 We showcase the versatility of TDN by generating terrains with human-provided sketches, in contrast to the dataset introduced in Sec. 4.1 of which the sketches are generated by algorithms.
Fig. \ref{fig:figure5} shows that TDN is responsive and capable of adapting to different user sketches whilst generating plausiable terrains. TDN can handle conflicting and complex user sketches. In the first example, the ridges overlaps with basins, peaks, and rivers. TDN shows successful and realistic generation. Especially when compared with a GAN-based method, GAN puts most of its weights on one sketch - the peaks of the terrain, and fails to take other inputs into consideration. The same finding can be verified through the second example.
TDN can integrate climatic conditioning, as in the third example. Given seemingly random input, TDN produces a more realistic terrain compare to GAN.
By synthesising these climatic conditions, TDN delivers an output that is both a faithful representation of the input and an authentic digital portrayal of real-world terrains. This demonstrates our model's unique ability to achieve both user customization and environmental realism.

\subsection{Limitations}
\par Several limitations are with TDN and further studies need to be conducted in the future. First, the quality of the terrain generation is overly reliant on the scale of the training data. The diffusion model learns the data's distribution and can only denoise samples within the given distribution. Consequently, attempts to encompass a wide range of terrain types come at the cost of sacrificing granular details. This is due to the challenging nature of obtaining high-resolution or ultra-high-resolution terrain data that contains a diverse array of terrain types. This might be rectified by having high quality data, for example, sourcing terrains from multiple sources, and hence building a large and comprehensive database. However, the sheer volume of the training data introduces a significant challenge, as the associated costs of training would be substantially high.
Second, the training of the terrain and sketch autoencoders is done in an independent manner. When these components are combined with the diffusion model, the overall training process becomes computationally expensive. Meanwhile, the diffusion process for inference on a \(144\times144\) terrain map takes around 11 seconds for 36 steps. It could potentially impede the model's accessibility and widespread adoption.
Finally, the current control is mainly sketch-based with a fixed number of user-provided sketches. However, it is more promising to integrate with flexible guidance such as: customized sketch categories rather than using all of them; and guidance in other modalities like text.

\section{Conclusion}

This study presents a novel diffusion-based method - Terrain Diffusion Network (TDN) with user-controlled sketch guidance. A multi-level denoising scheme with three synthesizers is proposed to formulate structural and fine-grained level terrain characteristics, including the climatic patterns.  
Moreover, to maximise the efficiency of TDN, we further introduce terrain and sketch latent spaces with pre-trained autoencoders that is aimed for user sketch consistency. TDN integrates this guidance directly, affording users superior conformity to their inputs. Comprehensive experiments on our newly collected dataset clearly demonstrate the state-of-the-art performance of TDN.

\bibliography{TDN} 


\end{document}